\pdfoutput=1

\documentclass[11pt]{article}

\usepackage[preprint]{acl}

\usepackage{times}
\usepackage{latexsym}

\usepackage[T1]{fontenc}

\usepackage[utf8]{inputenc}

\usepackage{microtype}

\usepackage{inconsolata}

\usepackage{graphicx}

\usepackage[utf8]{inputenc} 
\usepackage[T1]{fontenc}    
\usepackage{hyperref}       
\usepackage{url}            
\usepackage{booktabs}       
\usepackage{amsfonts}       
\usepackage{nicefrac}       
\usepackage{microtype}      
\usepackage{xcolor}         
\usepackage{subfigure}
\usepackage{amsmath}
\usepackage{amssymb}
\usepackage{mathtools}
\usepackage{amsthm}
\usepackage{multicol}
\usepackage{multirow}
\usepackage{tablefootnote}
\usepackage{pifont}
\usepackage{makecell}

\theoremstyle{plain}
\usepackage[ruled,longend,noline]{algorithm2e}

\SetCommentSty{mycommfont}

\newcommand{\name}{FR-Spec}
\title{FR-Spec: Accelerating Large-Vocabulary Language Models \\via Frequency-Ranked Speculative Sampling}

\author{
 \textbf{Weilin Zhao\textsuperscript{1}\thanks{\ \ indicates equal contribution.}},
 \textbf{Tengyu Pan\textsuperscript{1}$^*$},
 \textbf{Xu Han\textsuperscript{1}\thanks{\ \ indicates corresponding authors.}},
 \textbf{Yudi Zhang\textsuperscript{2}},
 \textbf{Ao Sun\textsuperscript{3}},
 \textbf{Yuxiang Huang\textsuperscript{1}}
\\
 \textbf{Kaihuo Zhang\textsuperscript{4}},
 \textbf{Weilun Zhao\textsuperscript{4}},
 \textbf{Yuxuan Li\textsuperscript{1}},
 \textbf{Jianyong Wang\textsuperscript{1}},
 \textbf{Zhiyuan Liu\textsuperscript{1}$^\dag$},
 \textbf{Maosong Sun\textsuperscript{1}}
\\
\\
\textsuperscript{1}{Tsinghua University, Beijing, China.}
\textsuperscript{2}{Harbin Institute of Technology, Harbin, China.}
\\
\textsuperscript{3}{Beijing University of Posts and Telecommunications, Beijing, China.}
\textsuperscript{4}{OpenBMB}
\\
{\tt \{zwl23,pty23\}@mails.tsinghua.edu.cn, \{han-xu,liuzy\}@tsinghua.edu.cn}
}

\begin{document}
\maketitle
\begin{abstract}

Speculative sampling has emerged as an important technique for accelerating the auto-regressive generation process of large language models (LLMs) by utilizing a draft-then-verify mechanism to produce multiple tokens per forward pass. While state-of-the-art speculative sampling methods use only a single layer and a language modeling (LM) head as the draft model to achieve impressive layer compression, their efficiency gains are substantially reduced for large-vocabulary LLMs, such as Llama-3-8B with a vocabulary of 128k tokens. To address this, we present \name, a frequency-ranked speculative sampling framework that optimizes draft candidate selection through vocabulary space compression. By constraining the draft search to a frequency-prioritized token subset, our method reduces LM Head computation overhead by 75\% while ensuring the equivalence of the final output distribution. Experiments across multiple datasets demonstrate an average of 1.12$\times$ speedup over the state-of-the-art speculative sampling method EAGLE-2. Code available at \url{https://github.com/thunlp/FR-Spec}.

\end{abstract}

\section{Introduction}

Large language models (LLMs) have revolutionized the field of artificial intelligence (AI), enabling a wide range of applications from conversational AI to complex reasoning tasks~~\cite{gpt3,chatgpt,deepseekr1}. Over time, driven by the need to improve tokenization efficiency and support multilingual capabilities and domain-specific terminologies, the standard vocabulary size of LLMs has grown significantly, from a vocabulary of 32k tokens used in Llama-2~\cite{llama2} to the much larger vocabularies adopted by recent mainstream models. Notable examples include Llama-3~\cite{llama3} with 128k vocabulary tokens, Qwen-2.5~\cite{qwen25} with 152k vocabulary tokens, and DeepSeek-V3~\cite{deepseekv3} with 129k vocabulary tokens. While larger vocabularies enhance model capabilities~\cite{takase2024large, tao2024scaling}, the side effect of a large vocabulary on the generation speed of LLMs remains unstudied.

\begin{figure}[t]
    \centering
    \includegraphics[width=0.47\textwidth]{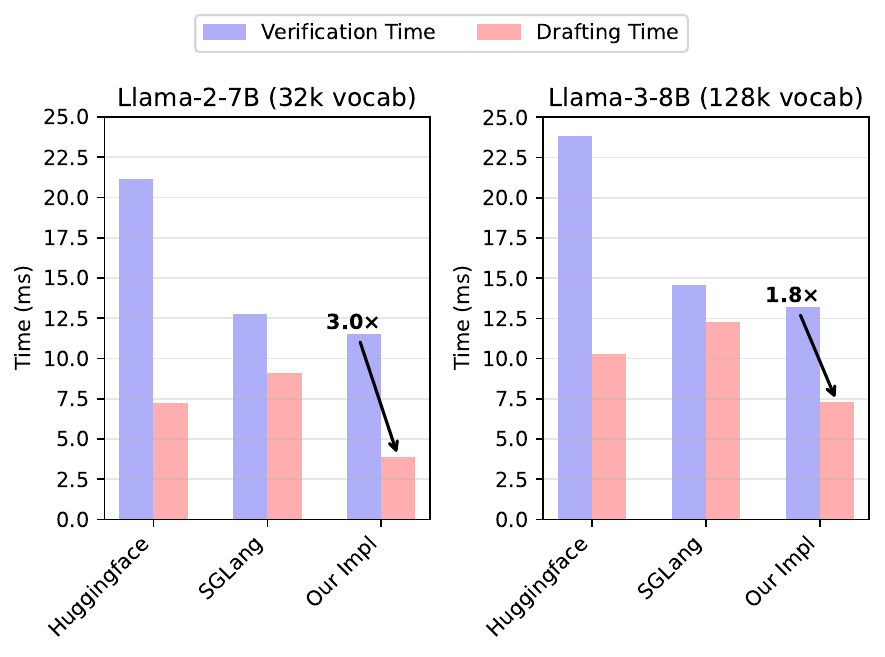}
    \caption{Comparison of the drafting and verification times of EAGLE-2 implemented by  three frameworks (Huggingface, SGLang, and our optimized implementation) for two vocabulary sizes: 32k (Llama-2-7B) and 128k (Llama-3-8B).}
    \label{fig:intro}
\end{figure}

To meet the demand for faster generation speed, speculative sampling~\cite{googlespec,deepmindspec} has emerged as a leading technique, particularly for deploying LLMs on resource-restricted devices such as PCs, laptops, and mobile phones.
These methods, such as Medusa~\cite{medusa} and EAGLE-2~\cite{eagle2}, employ a two-stage draft-then-verify mechanism. In each iteration, a lightweight draft model first predicts several draft sequences. Subsequently, the target LLM verifies the drafted tokens in parallel and accepts the longest correct subsequence matching the LLM's own predictions. This approach allows the LLM to validate multiple tokens in one forward pass. The recent state-of-the-art speculative sampling method, EAGLE-2, has made remarkable progress in reducing the time required for the drafting process, by employing an extremely lightweight architecture --- the drafting process relies solely on a single-layer transformer. Despite its simplicity, EAGLE-2 achieves impressive drafting quality, enabling accurate and efficient token predictions that significantly accelerate the overall generation process.

Although speculative sampling has shown promising results, its research highly relies on the Huggingface framework for experimental speedup evaluation. As a result, the negative effects of large vocabularies are obscured due to Python overhead, CPU processing, and suboptimal operator implementations. 
By implementing EAGLE-2 in native C and CUDA, we observed a substantial increase in drafting time when transitioning from small to large vocabulary models, as illustrated in Figure~\ref{fig:intro}.

\begin{figure}[t]
    \centering
    \includegraphics[width=0.49\textwidth]{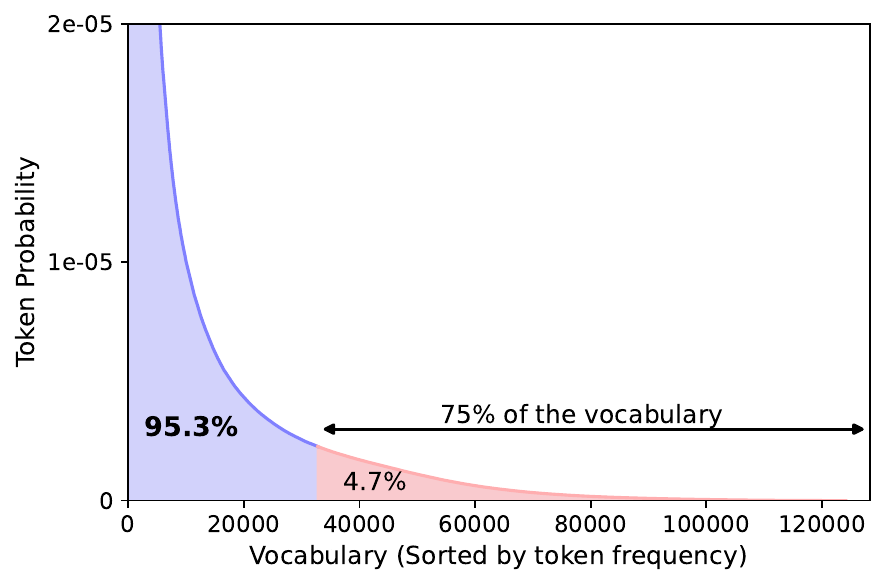}
    \caption{Token frequency distribution, statistically analyzed using the tokenizer of Llama-3-8B on a subset of 1B tokens randomly sampled from the SlimPajama-627B dataset~\cite{cerebras2023slimpajama}. As shown in the figure, 75\% of the vocabulary tokens account for less than 5\% of all token occurrences in the dataset, presenting a ``Long Tail'' effect.}
    \label{fig:longtail}
\end{figure}

To tackle this challenge and achieve further speedup, we introduce \name, a frequency-ranked speculative sampling framework that optimizes draft candidate selection through vocabulary space compression.
Our key inspiration is drawn from the long-tailed distribution~\cite{zipf-law} of token frequencies in natural languages, as depicted in Figure~\ref{fig:longtail}. This distribution indicates that a significant portion of tokens in the vocabulary of LLMs are rarely used. By restricting the draft search to a frequency-prioritized subset of high-probability tokens, we reduce the computational overhead of the language modeling (LM) Head by 75\%.
While this results in a slight reduction in drafting accuracy, it significantly improves the overall generation speed. Importantly, \name~preserves the mathematical equivalence of the verification process, ensuring that the final output distribution remains unaltered compared with the original sampling methods.

Our contributions are summarized as follows:

1. \textbf{A Systematic Time Breakdown of Speculative Sampling}. To address the current limitations where the bottleneck analyses of speculative sampling are either insufficiently explored or commonly rely on sub-optimized implementations (e.g. Huggingface Transformers), we develop a highly optimized implementation and conduct detailed profiling. Surprisingly, our analysis reveals that the \textbf{LM Head}, rather than the transformer layers, \textbf{is the primary bottleneck} in the drafting process.

2. \textbf{Frequency-Ranked Speculative Sampling}. To mitigate the computational cost of the LM Head, we propose a frequency-prioritized subset of the vocabulary for the drafting process, while retaining the full vocabulary for verification. Our method, \name, is designed as a plug-and-play solution, compatible with existing speculative sampling techniques and requiring no retraining. Our approach achieves an extra 1.12$\times$ speedup when integrated with the current state-of-the-art method EAGLE-2 and 1.08$\times$ speedup when integrated with Medusa.

\begin{figure*}[t]
    \centering
    \includegraphics[width=0.97\textwidth]{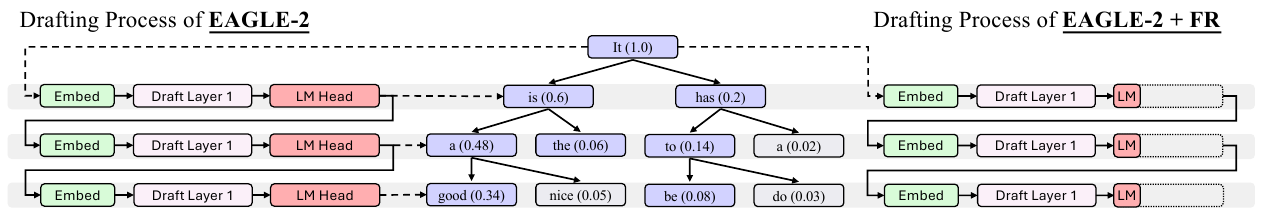}
    \caption{(Left) The drafting process of EAGLE-2 when prompt$~P=$``It'', beam $width =2$ and search $depth =3$. It picks out the top $K=8$ probability tokens (purple) as the draft tree. (Right) The drafting process of \name, where the LM Head is cropped during the drafting process while the beam search procedure remains the same.}
    \label{fig:framework_1}
\end{figure*}

\begin{figure}[t]
    \centering
    \includegraphics[width=0.47\textwidth]{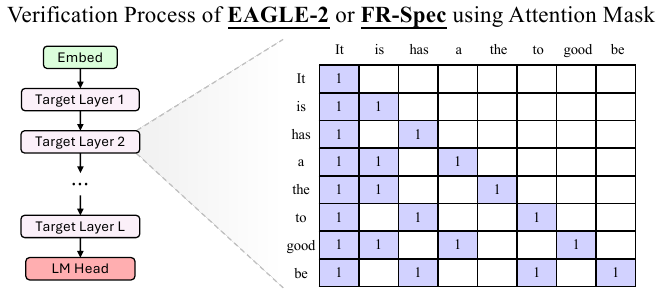}
    \caption{The illustration of the verification process for EAGLE-2 and \name, given the draft in Figure~\ref{fig:framework_1}. \name~solely modifies the drafting process while the verification process remains consistent with EAGLE-2.}
    \label{fig:framework_2}
\end{figure}

\section{Preliminary}

In this section, we introduce the concept of speculative sampling by taking the state-of-the-art method EAGLE-2~\cite{eagle2} as an example. The fundamental principles and operations of EAGLE-2 can serve as a representative model; other speculative sampling methods follow similar logic and can refer to the related work section (Section~\ref{sec:relate}).

An LLM $\mathcal{T}$ with the vocabulary $\mathcal{V}$ consists of an embedding layer $\mathcal{E}$, $L$ layers of transformer blocks $\mathcal{T}_{\text{layer}}^{(1)},\mathcal{T}_{\text{layer}}^{(2)},\cdots,\mathcal{T}_{\text{layer}}^{(L)}$, and an LM Head with the weight $\mathbf{W}_{\text{LM}}\in\mathbb{R}^{|\mathcal{V}|\times d}$. The embedding layer is responsible for mapping tokens $\mathbf{x}\in\mathbb{R}^{n}$ into a $d$-dimensional latent space. After using the transformer blocks to encode token embeddings, the LM Head projects the output representations back into the vocabulary space. Finally, a softmax function is applied to the vocabulary space to get output token probabilities.
Overall, the model $\mathcal{T}$ can be represented as first calculating the last hidden state $\mathbf{H}_{\mathcal{T}}(\mathbf{x}) \in \mathbb{R}^{n\times d}$, followed by the LM Head projection and softmax computation to obtain the final output token probabilities:
\begin{equation}
\begin{aligned}
\mathbf{H}_{\mathcal{T}}(\mathbf{x}) &= \mathcal{T}_{\text{layer}}^{(L)}\circ\cdots\circ\mathcal{T}_{\text{layer}}^{(2)}\circ \mathcal{T}_{\text{layer}}^{(1)}\big(\mathcal{E}(\mathbf{x})\big), \\
\mathcal{T}(\mathbf{x}) &= \text{Softmax}(\mathbf{H}_{\mathcal{T}}(\mathbf{x})~\mathbf{W}_{\text{LM}}^T).
\end{aligned}
\label{eq:target}
\end{equation}

For the LLM $\mathcal{T}$, EAGLE-2 trains a lightweight draft model $\mathcal{D}$ to approximate $\mathcal{T}$’s behavior while drastically reducing computational overhead. The draft model $\mathcal{D}$ is structured as a single-layer transformer, with its latent dimension $d$ being identical to that of the target LLM. For the draft model, the parameters of its embedding layer and LM head are directly sourced from the target LLM and are frozen during the training process. The transformer layer of the draft model is then trained on some training data to make the draft model mimic the generation results of the target LLM. To summarize, $\mathcal{D}$ can be represented as calculating the hidden state $\mathbf{H}_{\mathcal{D}}(\mathbf{x}) \in \mathbb{R}^{n\times d}$, and conducting LM Head projection:
\begin{equation}
\begin{aligned}
\mathbf{H}_{\mathcal{D}}(\mathbf{x}) &= \mathcal{D}_{\text{layer}}^{(1)}(\mathcal{E}(\mathbf{x})), \\
\mathcal{D}(\mathbf{x}) &= \text{Softmax}(\mathbf{H}_{\mathcal{D}}(\mathbf{x})~\mathbf{W}_{\text{LM}}^T).
\end{aligned}
\label{eq:draft}
\end{equation}
EAGLE-2 actually combines $\mathbf{H}_{\mathcal{T}}(\mathbf{x})$ from the target LLM with $\mathcal{E}(\mathbf{x})$ on the draft input, but this does not affect the presentation of our paper, so the formula is simplified as Eq.(\ref{eq:draft}) for clarity.

During inference, given a specific prompt $P$, EAGLE-2 adopts a beam-search algorithm based on the softmax output of the draft model to complete a drafting process. Given a beam width and a search depth, EAGLE-2 uses the draft model $\mathcal{D}$ to forward $depth$ times and then select the top $K$ probability tokens from the beam search history as the draft. As illustrated in Figure~\ref{fig:framework_1} (left), EAGLE-2 finally generates a draft tree consisting of multiple draft sequences, and the draft tree is then verified by the target LLM $\mathcal{T}$ using a tree attention mask demonstrated in Figure~\ref{fig:framework_2}. 
The special tree attention mask is created based on the draft tree, where each token can only see tokens in its ancestral path and in the prompt prefix $P$.

\section{Methodology}

\subsection{Identifying Key Bottlenecks for Speculative Sampling}

To gain deeper insights into the time breakdown of speculative sampling and quantify the contribution of each component, we first implement an optimized speculative sampling framework and employ profiling tools to analyze the key bottlenecks of EAGLE-2 under our optimized framework.

\begin{figure}[t]
    \centering
    \includegraphics[width=0.47\textwidth]{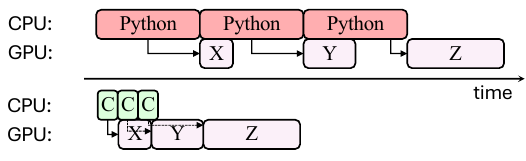}
    \caption{Comparison of Python-based implementation and C-based implementation. X, Y, and Z represent three different short-duration computational tasks. }
    \label{fig:cimpl}
\end{figure}

\textit{\textbf{Filtering out Non-Algorithmic Overheads}}.
Before conducting the analysis, it is crucial to rule out the analysis errors caused by sub-optimized framework implementations. For instance, despite its widespread use and convenience, Python's dynamic typing and interpreted nature can introduce inefficiencies that are not directly related to the analyzed algorithms. For example, the beam search algorithm in EAGLE-2, characterized by a large number of short-duration computational tasks, leads to significant latency issues in the original PyTorch~\cite{paszke2019pytorch} implementation, as illustrated in Figure~\ref{fig:cimpl}. Specifically, executing these tasks requires frequent waiting for Python's launch commands, making them one of the bottlenecks. To mitigate this, we reimplement EAGLE-2 using native C and CUDA and preallocate all required memory in advance. This eliminates the overhead associated with Python's interpreter. As demonstrated in Figure~\ref{fig:cimpl}, this optimization can significantly reduce latency and make the overall execution more streamlined.

Additionally, suboptimal operator implementations can introduce significant implementation-level overheads.
We thus modify FlashAttention~\cite{flashattention-2} to support complex tree attention masks as in Figure~\ref{fig:framework_2}. To minimize the performance impact of memory access for attention masks, we optimize the process by transmitting only the portion of the mask corresponding to the draft tokens, given that the prompt prefix $P$ is entirely causal. 
Moreover, since EAGLE-2 (and other speculative sampling methods) typically involves no more than 64 draft tokens, we employ bitmask compression using ``uint64'' to ensure more contiguous and compact memory access patterns, thereby enhancing overall efficiency.

\begin{figure}[t]
    \centering
    \includegraphics[width=0.47\textwidth]{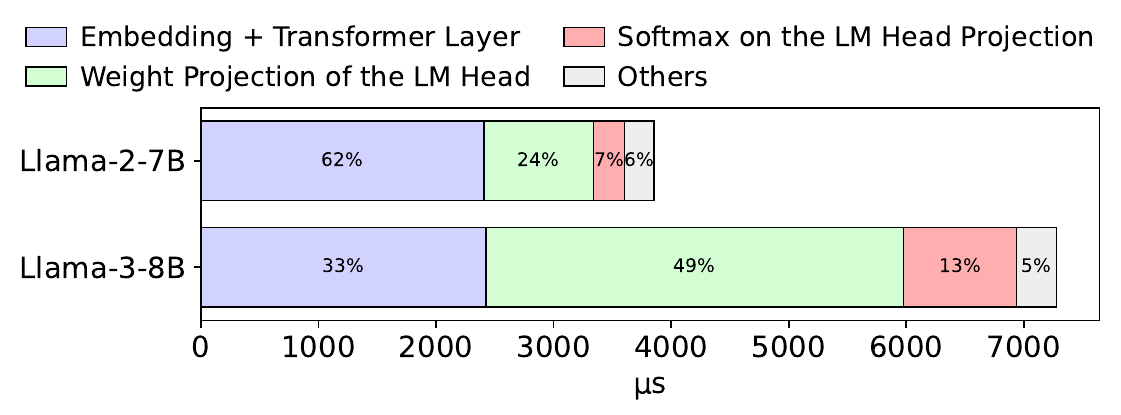}
    \caption{Time breakdown of the drafting process of EAGLE-2. We profile the EAGLE-2 trained on Llama-2-7B (32k vocabulary) and the EAGLE-2 trained on Llama-3-8B (128k vocabulary).}
    \label{fig:profile}
\end{figure}

\textit{\textbf{Wall Time Breakdown}}.
Based on our optimized implementation framework, we observe a substantial increase in drafting time when shifting from small vocabulary LLMs to large vocabulary LLMs, as in Figure~\ref{fig:intro}.
To investigate the underlying reasons for this, we conduct a comprehensive profiling analysis on our proposed framework. As shown in Figure~\ref{fig:profile}, the computational bottleneck in the drafting process has shifted from the transformer layer, which is traditionally considered time-consuming, to the LM Head. The vocabulary size directly causes such a significant disparity associated with the LM Head component. 
Additionally, the softmax function, which operates across the dimension of the vocabulary size, also exhibits a notable increase in wall time.

Specifically, the profiling data indicates that the LM Head component accounts for a substantial 49\% of the total computational time in the drafting process, nearly half of the entire processing time. When accounting for the combined computation time of the LM Head and the softmax operation, both directly proportional to the vocabulary size, the proportion increases to 62\%. In contrast, the transformer layer accounts for only 33\% of the computation time. This indicates that vocabulary-related computations require nearly twice the time of the transformer layer's operations.

These findings indicate that while a large vocabulary has a relatively moderate impact on the speed of the LLM itself, the scenario shifts significantly within the speculative sampling framework. This is due to the highly lightweight architecture of the drafting model, which follows a 1:1 ratio of one transformer layer to one LM Head. This underscores the importance of optimizing vocabulary-related operations to enhance the efficiency of speculative sampling in large vocabulary settings.

\subsection{Addressing the Bottleneck Caused by Large Vocabulary}

To optimize for large-vocabulary scenarios, we conducted a corpus-level token-frequency analysis, which revealed that the vast majority of tokens hardly appear in the corpus, demonstrating a sparse pattern across the vocabulary.
We then utilize the sparse pattern to let the draft model focus exclusively on drafting high-probability tokens, while tokens with extremely low probabilities of occurrence are left to be handled by the LLM.

\textit{\textbf{Corpus-Level Token Statistics}}.
\label{sec:frequency-pattern}
We begin by analyzing the token frequency distribution across a pretraining corpus SlimPajama-627B~\cite{cerebras2023slimpajama}. The data in the pre-training corpus encompasses a vast amount of information from diverse fields. It is highly suitable for token-frequency analysis. As illustrated in Figure~\ref{fig:longtail}, we use a 1 billion token subset of the pretraining corpus to get the corpus-level token statistics. Our statistical study reveals a pronounced long-tail pattern: a small subset (25\%) of tokens (e.g., common words, punctuation, and general-domain terms) accounts for the majority of occurrences (95\%), while the remaining (75\%) tokens exhibit sparse frequencies (5\%). This observation motivates our core design: restricting the draft model's generation scope to the small subset of high-frequency tokens can significantly accelerate the drafting process without sacrificing much draft quality.

\textit{\textbf{\name}}.
We propose a frequency-ranked drafting mechanism. Let $\mathcal{V}$ denote the full vocabulary of the language model, and $\mathcal{V}_{\text{high}} \subset \mathcal{V}$ represent the subset of high-frequency tokens identified through previously mentioned corpus-level statistics. At each generation step, instead of computing probabilities over the entire vocabulary, we restrict the drafting model's output distribution $\mathcal{D}(\mathbf{x})$ to $\mathcal{V}_{\text{high}}$, as shown in Figure~\ref{fig:framework_1} (right). We only limit the vocabulary of the drafting process while keeping the verification process untouched.

To this end, we first create a sub matrix $\mathbf{\tilde W}_{\text{LM}} \in\mathbb{R}^{|\mathcal{V_{\text{high}}}|\times d}$ from $\mathbf{W}_{\text{LM}}\in\mathbb{R}^{|\mathcal{V}|\times d}$, by letting
\begin{equation}
\mathbf{\tilde W}_{\text{LM}}[i, :] = \mathbf{W}_{\text{LM}}[\mathcal{V}_{\text{high}}[i], :], i=1\dots|\mathcal{V}_{\text{high}}|.
\end{equation}
Then we change the draft equation from Eq.(\ref{eq:draft}) to 
\begin{equation}
\mathcal{D}_{\text{FR}}(\mathbf{x}) = \text{Softmax}(\mathbf{H}_{\mathcal{D}}(\mathbf{x})~\mathbf{\tilde W}_{\text{LM}}^T)
\label{eq:draft_fr}
\end{equation}
As can be seen, from changing Eq.(\ref{eq:draft}) to Eq.(\ref{eq:draft_fr}), the computational complexity of the LM Head projection is reduced from the original $\mathrm{O}(n d |\mathcal{V}|)$ to $\mathrm{O}(n d |\mathcal{V}_{\text{high}}|)$, achieving a reduction by a factor of $\frac{|\mathcal{V}|}{|\mathcal{V}_{\text{high}}|}$. Additionally, the input dimension of Softmax is reduced from $\mathbf{H}_{\mathcal{D}}(\mathbf{x}) \mathbf{W}_{\text{LM}}^T \in \mathbb{R}^{n \times |\mathcal{V}|}$ to $\mathbf{H}_{\mathcal{D}}(\mathbf{x}) \mathbf{\tilde{W}}_{\text{LM}}^T \in \mathbb{R}^{n \times |\mathcal{V}_{\text{high}}|}$. The operation time of the softmax function, proportional to the input size, is also decreased by a factor of $\frac{|\mathcal{V}|}{|\mathcal{V}_{\text{high}}|}$ when using a reduced vocabulary subset.

By using a small subset of the original vocabulary, \name~indicates a context-related acceleration paradigm: sequences dominated by high-frequency tokens benefit from reduced computational overheads. While those regions requiring low-frequency tokens (e.g., rare named entities or technical terms) inherently bypass acceleration. We will balance this tradeoff in the subsequent experiment section and demonstrate that \textbf{the benefits of our approach outweigh its drawbacks}.

\begin{table*}[t]
\centering
\tabcolsep=1em
\scalebox{0.80}{
\begin{tabular}{l|ccccccc|c}
\toprule
Configuration & \textbf{MT.} & \textbf{Conv.} & \textbf{RAG} & \textbf{Math} & \textbf{QA} & \textbf{Summ.} & \textbf{Code} & \textbf{Average} \\
\midrule
Full Vocab (128k) & 3.66 & 4.12 & 4.05 & 4.29 & 3.49 & 3.68 & 3.92 & 3.89 (100\%)\\
\midrule
+FR 64k (ShareGPT) & 3.45 & 4.08 & 3.89 & 4.20 & 3.40 & 3.56 & 3.83 & 3.77 (96.9\%)\\
+FR 32k (ShareGPT)& 3.23 & 3.95 & 3.59 & 4.04 & 3.25 & 3.31 & 3.62 & 3.57 (91.8\%)\\
+FR 16k (ShareGPT) & 3.03 & 3.71 & 3.30 & 3.74 & 3.04 & 3.02 & 3.40 & 3.32 (85.3\%)\\
+FR 8k (ShareGPT) & 2.82 & 3.42 & 2.95 & 3.45 & 2.82 & 2.77 & 3.19 & 3.06 (78.7\%)\\
\midrule
+FR 64k (SlimPajama) & 3.59 & 4.07 & 3.98 & 4.26 & 3.42 & 3.65 & 3.62 & 3.80 (97.7\%)\\
+FR 32k (SlimPajama) & 3.39 & 3.89 & 3.85 & 4.15 & 3.34 & 3.51 & 3.29 & 3.63 (93.3\%)\\
+FR 16k (SlimPajama) & 3.20 & 3.63 & 3.56 & 3.84 & 3.19 & 3.28 & 3.10 & 3.40 (87.4\%)\\
+FR 8k (SlimPajama) & 2.98 & 3.33 & 3.25 & 3.52 & 2.97 & 2.98 & 2.86 & 3.13 (80.5\%)\\
\bottomrule
\end{tabular}
}
\caption{Average accepted length for Llama-3-8B under different \name~configurations. The numbers in parentheses (97.7\%) indicate the ratio achieved compared to the full vocabulary baseline.}
\label{tab:acceptance_llama3}
\end{table*}

\begin{table*}[h]
\centering
\tabcolsep=1em
\scalebox{0.80}{
\begin{tabular}{l|ccccccc|c}
\toprule
Method & \textbf{MT.} & \textbf{Conv.} & \textbf{RAG} & \textbf{Math} & \textbf{QA} & \textbf{Summ.} & \textbf{Code} & \textbf{Average} \\
\midrule
Vanilla & 90.94 & 90.43 & 83.43 & 91.16 & 91.05 & 86.63 & 90.10 & 89.11 (1.00$\times$) \\
\midrule
EAGLE-2 & 176.79 & 203.41 & 168.05 & 209.88 & 166.60 & 167.12 & 175.11 & 180.99 (2.03$\times$) \\
\midrule
+FR 64k & 192.85 & 224.52 & 178.53 & 231.99 & 183.17 & 183.86 & 183.11 & 196.86 (2.21$\times$) \\
\textbf{+FR 32k} & \textbf{195.60} & \textbf{227.68} & \textbf{184.85} & \textbf{243.36} & \textbf{190.27} & \textbf{188.14} & \textbf{183.19} & \textbf{201.87 (2.27$\times$)} \\
+FR 16k & 194.02 & 223.32 & 178.22 & 233.69 & 188.60 & 182.26 & 176.70 & 196.69 (2.21$\times$) \\
+FR 8k & 185.78 & 210.66 & 167.64 & 218.57 & 180.40 & 170.97 & 167.85 & 185.98 (2.09$\times$) \\
\bottomrule
\end{tabular}
}
\caption{Decoding speed (token/s) of \name~and baselines on Llama-3-8B under our implementation framework using temperature=0. The numbers in parentheses (2.27$\times$) indicate the speedup compared to the baseline (Vanilla).}
\label{tab:speed_llama3_8b_temp0}
\end{table*}

\section{Experiments}

This section focuses on evaluating \name~on various tasks when applying to various large-vocabulary LLMs to demonstrate the efficiency and effectiveness of \name.

\subsection{Experimental Settings}

\textbf{Datasets}. To comprehensively assess the speed performance of various speculative sampling methods, we evaluate \name~across seven representative text generation tasks: machine translation (MT.), multi-turn conversation (Conv.), retrieval-augmented generation (RAG), arithmetic reasoning (Math), question answering (QA), document summarization (Summ.), and code generation (Code). Specifically, we adopt Spec-Bench~\cite{specbench} benchmark, a widely used benchmark for speculative sampling, which covers the first six subtasks, with datasets drawn from the following sources: Translation from WMT14 DE-EN~\cite{wmt14}, Multi-turn Conversation from MT-bench~\cite{mtbench}, RAG and QA from Natural Questions~\cite{natural_questions}, Math from GSM8K~\cite{gsm8k}, and Summarization from CNN/Daily Mail~\cite{cnn/daily_mail}, with 80 entries per subtask. In addition, we include the HumanEval~\cite{humaneval} benchmark, which focuses on code generation tasks and contains 164 entries. Following \citet{specbench}, we set the maximum generation lengths to 1024 for all subtasks in Spec-Bench and 512 for HumanEval.

\textbf{Models}. We select Llama-3-8B-Instruct (128k vocabulary)~\cite{llama3}, Llama-3.2-1B-Instruct (128k vocabulary) and Qwen-2-7B-Instruct (152k vocabulary)~\cite{qwen2} as the language models for experiments. These models are recently representative and popular LLMs.

\textbf{Evaluation Methods}. We select vanilla autoregressive decoding and EAGLE-2 as our baselines. We integrate \name~with EAGLE-2, which we called ``EAGLE-2 (+FR)'' later. We report the mean acceptance length and decoding speed (token/s). Following the settings in Spec-Bench~\cite{specbench}, we set the search depth of EAGLE-2 to 6 and the total amount of draft tokens to 60.

\textbf{Hardware Settings}. Experiments in this section are performed on 1 $\times$ NVIDIA 80GB A800 GPU. The CPU used is the Intel(R) Xeon(R) Platinum 8470. Experiments on other platforms can refer to Appendix~\ref{appendix:llama3-1b}.

\subsection{Accept Length}
\label{sec:accept_length}

To thoroughly investigate the impact of the frequency-ranked drafting mechanism on existing speculative sampling frameworks, we conducted experiments across seven subtasks, measuring the average acceptance length under different vocabulary truncation strategies.
The average acceptance length is an important metric in speculative sampling. It quantifies the number of draft tokens that are verified as correct in each iteration. It serves as an effective assessment of drafting quality and is one important factor that affects the final speedup aside from the drafting time.

Specifically, we tried two datasets for token frequency statistics: (1) SlimPajama-627B~\cite{cerebras2023slimpajama}. We sample a 1 billion token subset from it. Conducting tokenization on this subset requires less than 30 minutes. (2) ShareGPT~\cite{sharegpt2023}. ShareGPT is the training data for EAGLE-2, and we use the whole dataset, which comprises 134 million tokens.

Based on the token-frequency statistics, we select four different vocabulary sizes ($|\mathcal{V}_{\text{high}}|=\{8\text{k},16\text{k},32\text{k},\text{64k}\}$) to serve as the new LM Head configurations for the draft model.
Table~\ref{tab:acceptance_llama3} reports the average acceptance length of the Llama-3-8B model across different \name~configurations. As shown in the results, when the vocabulary size of the draft model was halved from 128k to 64k, the average acceptance length only decreased slightly (2.3\% for SlimPajama and 3.1\% for ShareGPT). 
This result is consistent with the ``long-tail'' characteristic of token frequency analyzed in Section~\ref{sec:frequency-pattern}. When the vocabulary size was reduced to 8k, a significant shortening of the acceptance length was observed. This finding underscores the need to strike a balance between the draft accuracy and drafting time of the draft model. In Section~\ref{sec:exp_speed} below, we will conduct an in-depth analysis of this trade-off, taking into account the drafting time.

Notably, frequency statistics derived from SlimPajama outperform those from ShareGPT in terms of average accept length. The observation remains consistent when applied to Qwen-2-7B and Llama-3.2-1B, as detailed in Appendix~\ref{appendix:qwen2-accept-length} and~\ref{appendix:llama3-1b}. We attribute this difference to the higher quality and the larger volume of the SlimPajama data. Therefore, we adopted SlimPajama-based statistics for subsequent experiments.

\subsection{Decoding Speed}
\label{sec:exp_speed}

Based on our native C and CUDA implementation, we evaluate the speed of the proposed \name~method and baselines on the Llama-3-8B model, as detailed in Table \ref{tab:speed_llama3_8b_temp0}.
As can be seen, \name~surpasses the original EAGLE-2 in all vocabulary configurations.
Comparing different vocabulary sizes, setting $|\mathcal{V}_{\text{high}}|=32\text{k}$ consistently outperforms other vocabulary configurations. Specifically, this configuration achieves an average speedup improvement of 11.8\% over EAGLE-2, achieving the best trade-off between draft accuracy and drafting time. Experiments on Llama-3-1B can refer to Appendix~\ref{appendix:llama3-1b}, where \name~achieves 24.2\% extra speedup over EAGLE-2.

\begin{figure}[t]
    \centering
    \includegraphics[width=0.44\textwidth]{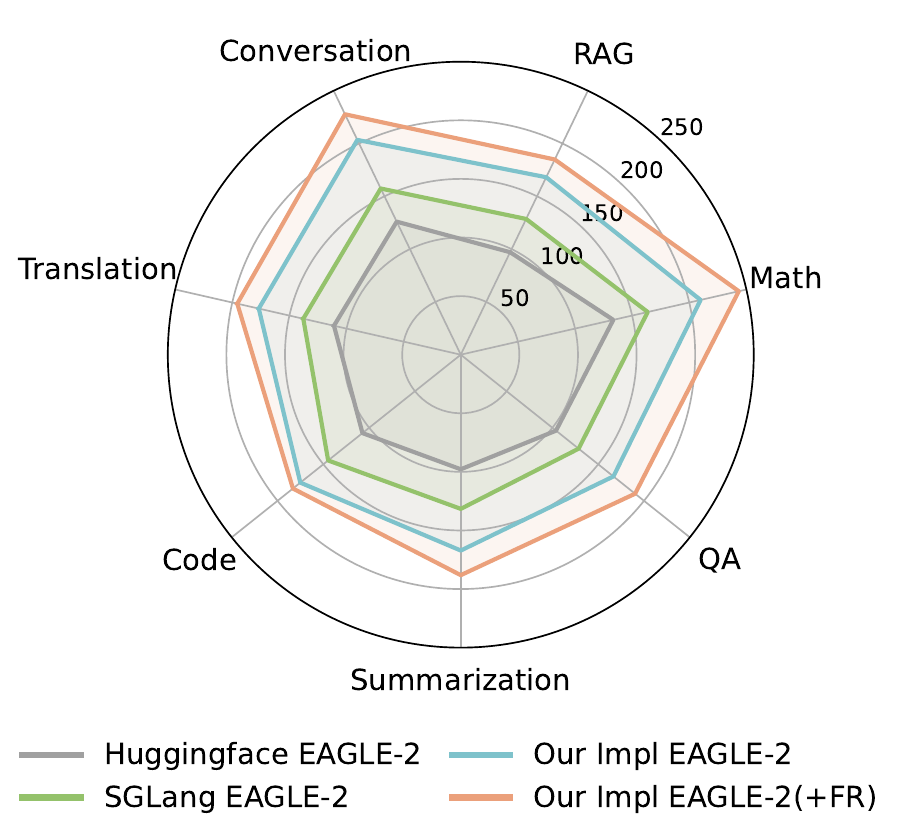}
    \caption{Decoding speed (token/s) of \name~and EAGLE-2 for Llama-3-8B under different frameworks.}
    \label{fig:speed_compare}
\end{figure}

Furthermore, we conducted speed analyses between our implementation and mainstream frameworks, namely Huggingface and SGLang. As the experimental results demonstrated in Figure~\ref{fig:speed_compare}, our optimized EAGLE-2 achieves average speedups of 1.63$\times$ and 1.28$\times$ compared to the original HuggingFace and SGLang versions, respectively. The \name~further improves these performance gains, with speedups of 1.82$\times$ and 1.42$\times$ over the HuggingFace and SGLang implemented EAGLE-2, respectively.

\name~supports both greedy decoding and random sampling. As illustrated in Table~\ref{tab:speed_llama3_8b_temp1}, \name~can achieve a speedup ratio of 1.13$\times$ compared to EAGLE-2 at a temperature of 1. This performance is comparable to the acceleration observed at the temperature of 0, showing that \name~is effective at different generation settings.

\subsection{Model Performance}

To validate the correctness of our \name, we assessed the generation quality of the Llama-3-8B model across two tasks: code generation using the HumanEval benchmark and mathematical reasoning with the GSM8K benchmark. We compare the model’s performance between the HuggingFace implementation and our optimized implementation in Table~\ref{tab:correctness}, in both greedy decoding (temperature=0) and random sampling (temperature=1) scenarios.

\begin{table}[t]
\centering
\setlength{\tabcolsep}{4pt}
\scalebox{0.74}{
\begin{tabular}{l|c|cc|cc}
\toprule
\multirow{2}{*}{Benchmark} & Vanilla & \multicolumn{2}{c|}{EAGLE-2} & \multicolumn{2}{c}{EAGLE-2(+FR 32k)} \\
& token/s & token/s & Speedup & token/s & Speedup \\
\midrule
MT. & 90.32 & 171.03 & 1.89$\times$ & \textbf{188.69} & \textbf{2.09}$\times$ \\
Conv. & 89.85 & 187.95 & 2.09$\times$ & \textbf{212.08} & \textbf{2.36}$\times$ \\
RAG & 83.18 & 159.37 & 1.92$\times$ & \textbf{178.64} & \textbf{2.15}$\times$ \\
Math & 89.75 & 196.34 & 2.19$\times$ & \textbf{237.96} & \textbf{2.65}$\times$ \\
QA & 90.58 & 155.10 & 1.71$\times$ & \textbf{182.59} & \textbf{2.02}$\times$ \\
Summ. & 87.41 & 158.72 & 1.82$\times$ & \textbf{182.70} & \textbf{2.09}$\times$ \\
Code & 89.77 & 180.67 & 2.01$\times$ & \textbf{183.54} & \textbf{2.04}$\times$ \\
\midrule
Average & 88.69 & 172.74 & 1.95$\times$ & \textbf{195.17} & \textbf{2.20}$\times$ \\
\bottomrule
\end{tabular}
}
\caption{Decoding speed (token/s) of Llama-3-8B using temperature=1 under our implementation.}
\label{tab:speed_llama3_8b_temp1}
\end{table}

\begin{table}[t]
\centering
\setlength{\tabcolsep}{4pt}
\scalebox{0.74}{
\begin{tabular}{lc|ll|ll}
\toprule
\multicolumn{2}{c|}{} & \multicolumn{2}{c|}{Huggingface} & \multicolumn{2}{c}{Our Implementation} \\
Benchmark & Temp & Vanilla & EAGLE-2 & Vanilla & \name \\
\midrule
\multirow{2}{*}{HumanEval} 
& 0 & 54.9 & 54.9 & 57.3 & 58.5 \\
& 1 & 51.0\small$\pm$1.4 & 50.6\small$\pm$3.1 & 51.1\small$\pm$1.2 & 51.2\small$\pm$1.2 \\
\cmidrule{1-6}
\multirow{2}{*}{GSM8K}
& 0 & 76.8 & 77.0 & 76.3 & 76.1 \\
& 1 & 70.8\small$\pm$2.0 & 66.5\small$\pm$2.9 & 65.6\small$\pm$1.8 & 67.1\small$\pm$0.8 \\

\bottomrule
\end{tabular}
}
\caption{Performance of the Llama-3-8B model on math reasoning and code generation tasks across two implementation frameworks. Due to variability in results with temperature=1, we report the average scores and variance across five different random seeds.}
\label{tab:correctness}
\end{table}

The experimental results indicate that the performance across both implementations is comparable, with only minor discrepancies. These differences are expected, as different implementations use different computational orders, resulting in floating-point numerical errors that accumulate within the model layers.

\subsection{Integration to Other Speculative Methods}

As a plug-and-play acceleration solution that is compatible with various speculative sampling methods, we further assess \name~by integrating \name~to Medusa~\cite{medusa}, another representative speculative sampling method. Table~\ref{tab:speed_llama3_8b_medusa} presents the performance of \name~in our optimized implementation of Medusa, where \name~achieve 1.08$\times$ extra speedup. The experimental results demonstrate that while our previous analysis primarily focused on EAGLE-2, our method also shows effectiveness when applied to other representative speculative sampling approaches, exhibiting strong compatibility and user-friendliness across different implementations.

\begin{table}[h]
\centering
\setlength{\tabcolsep}{4pt}
\scalebox{0.76}{
\begin{tabular}{l|c|cc|cc}
\toprule
\multirow{2}{*}{Benchmark} & Vanilla & \multicolumn{2}{c|}{Medusa} & \multicolumn{2}{c}{Medusa (+FR 32k)} \\
& token/s & token/s & Speedup & token/s & Speedup \\
\midrule
MT. & 90.94 & 146.42 & 1.61$\times$ & \textbf{157.54} & \textbf{1.73}$\times$ \\
Conv. & 90.43 & 157.99 & 1.75$\times$ & \textbf{169.26} & \textbf{1.87}$\times$ \\
RAG & 83.43 & 130.56 & 1.56$\times$ & \textbf{139.62} & \textbf{1.67}$\times$ \\
Math & 91.16 & 160.95 & 1.77$\times$ & \textbf{174.56} & \textbf{1.91}$\times$ \\
QA & 91.05 & 138.92 & 1.53$\times$ & \textbf{151.07} & \textbf{1.66}$\times$ \\
Summ. & 86.63 & 130.08 & 1.50$\times$ & \textbf{141.39} & \textbf{1.63}$\times$ \\
Code & 90.10 & 152.57 & 1.69$\times$ & \textbf{161.28} & \textbf{1.79}$\times$ \\
\midrule
Average & 89.11 & 145.36 & 1.63$\times$ & \textbf{156.39} & \textbf{1.76}$\times$ \\
\bottomrule
\end{tabular}
}
\caption{Decoding speed (token/s) of Llama-3-8B using temperature=0 under our implemented Medusa.}
\label{tab:speed_llama3_8b_medusa}
\end{table}

\section{Related Work}
\label{sec:relate}

This section mainly introduces model acceleration methods related to large vocabulary and speculative sampling. More details on how LLMs work can refer to surveys~\citep{qiu2020pre,han2021pre,bommasani2021opportunities,zhao2023survey}. Other acceleration methods such as quantization and distillation can refer to suverys~\citep{xu2023survey,li2024large}.

\subsection{Acceration on Large Vocabulary}
Recent advancements in large language models (LLMs) have prompted a growing interest in addressing the challenges associated with large vocabularies. While several optimization efforts have been proposed to tackle these issues, the majority focus primarily on the training phase.
Computing the LM Head and the loss function over large vocabularies requires storing a huge intermediate state before gradient computation. Therefore, MST~\cite{luo2024mini} and CCE~\cite{wijmans2024cut} tried to mitigate the memory overhead caused by computing loss functions over large vocabularies. These approaches address the issue by using input partitioning or weight partitioning, and conduct activation recomputation~\cite{chen2016training} during the backward propagation.
In addition to the aforementioned works that require no modifications to the model architecture, \citet{joulin2017efficient} proposes a hierarchical vocabulary structure to eliminate the computation of irrelevant vocabulary adaptively.

Constrained Decoding~\cite{hokamp2017lexically, dong2024xgrammar} restricts the vocabulary space to generate highly structured outputs, particularly in the context of LLM agents, where the generated content must adhere to specific formats, such as producing parsable code or invocable functions.

\subsection{Speculative Sampling}

Traditional autoregressive generation in LLMs suffers from low generation speed due to the sequential nature of token prediction. To address this limitation, speculative sampling has emerged as a promising approach, leveraging draft-then-verification paradigms to accelerate decoding~\cite{microsoftspec, googlespec, deepmindspec}. Existing speculative sampling methods can be categorized into two branches:
(1) \textit{retrieval-based drafting} approaches like PLD~\cite{pld}, LLMA~\cite{llma}, and REST~\cite{rest} retrieve relevant context from the prompt, gaining significant speedups in context-dependent tasks (e.g., summarization) by reusing retrieved text spans from the prompt.
(2) \textit{model-based drafting} methods exemplified by SpecInfer~\cite{specinfer}, DistillSpec~\cite{distillspec}, Medusa~\cite{medusa} and EAGLE~\cite{eagle2}, which employ a draft model for general-purpose acceleration. Our work focuses on the latter category due to its broader applicability.
The draft models' structures also differ. For example, Medusa generates draft tokens based solely on the model's last hidden state, using a ``MLP+LM Head'' structure, while EAGLE incorporates both the last hidden state and preceding tokens, using a transformer structure. Among these model-based drafting methods, EAGLE-2~\cite{eagle2} achieves the current state-of-the-art speed.

To further accelerate existing speculative sampling methods, recent advancements have been made at both the algorithmic and implementation levels.
At the algorithm level, HASS~\cite{zhang2025learning} has enhanced the training tasks for draft models, AdaEAGLE~\cite{zhang2024adaeagle} and OPT-Tree~\cite{wang2024opt} introduced adaptive draft tree structures at inference time. Additionally, TriForce~\cite{triforce} employs KV-Cache compression on draft models to accelerate the drafting process in long-context scenarios, Ouroboros~\cite{zhao-etal-2024-ouroboros} utilize Lookahead Decoding~\cite{fu2024break} to accelerates the draft models when the draft model is not lightweight enough.
From an implementation perspective, efficient LLM frameworks like vLLM~\cite{kwon2023efficient} and SGLang~\cite{zheng2024sglang} have integrated speculative sampling. DeFT~\cite{yao2025deft} leverages FlashAttention~\cite{flashattention-2} to enhance the efficiency of speculative sampling.

\section{Conclusion}

In this paper, we systematically analyze the overlooked issue of LM Head in speculative sampling.
Based on our frequency statistics, we propose a frequency-ranked optimization strategy to optimize the drafting process. 
We restrict the drafting space to a high-frequency subset of the vocabulary to make draft models faster.
Experiments demonstrate that by building on top of EAGLE-2 and Medusa, we can further achieve speedup ratios of 1.12$\times$ and 1.08$\times$, respectively.
\name~can be applied to most existing speculative sampling methods with one-click modification and requires no retraining.

\section*{Limitations}

Our current approach relies on static frequency analysis of the vocabulary, which, while effective, lacks adaptive mechanisms. Despite this limitation, the proposed solution has demonstrated promising compatibility. 
In the future, we will explore better dynamic mechanisms for further speedup.

\bibliography{custom}

\begin{thebibliography}{53}
\providecommand{\natexlab}[1]{#1}

\bibitem[{Bojar et~al.(2014)Bojar, Buck, Federmann, Haddow, Koehn, Leveling,
  Monz, Pecina, Post, Saint-Amand, Soricut, Specia, and Tamchyna}]{wmt14}
Ond{\v{r}}ej Bojar, Christian Buck, Christian Federmann, Barry Haddow, Philipp
  Koehn, Johannes Leveling, Christof Monz, Pavel Pecina, Matt Post, Herve
  Saint-Amand, Radu Soricut, Lucia Specia, and Ale{\v{s}} Tamchyna. 2014.
\newblock Findings of the 2014 workshop on statistical machine translation.
\newblock In \emph{Proceedings of the Ninth Workshop on Statistical Machine
  Translation}, pages 12--58.

\bibitem[{Bommasani et~al.(2021)Bommasani, Hudson, Adeli, Altman, Arora, von
  Arx, Bernstein, Bohg, Bosselut, Brunskill
  et~al.}]{bommasani2021opportunities}
Rishi Bommasani, Drew~A Hudson, Ehsan Adeli, Russ Altman, Simran Arora, Sydney
  von Arx, Michael~S Bernstein, Jeannette Bohg, Antoine Bosselut, Emma
  Brunskill, et~al. 2021.
\newblock On the opportunities and risks of foundation models.
\newblock \emph{arXiv preprint arXiv:2108.07258}.

\bibitem[{Brown et~al.(2020)Brown, Mann, Ryder, Subbiah, Kaplan, Dhariwal,
  Neelakantan, Shyam, Sastry, Askell et~al.}]{gpt3}
Tom Brown, Benjamin Mann, Nick Ryder, Melanie Subbiah, Jared~D Kaplan, Prafulla
  Dhariwal, Arvind Neelakantan, Pranav Shyam, Girish Sastry, Amanda Askell,
  et~al. 2020.
\newblock Language models are few-shot learners.
\newblock In \emph{Proceedings of NeurIPS}, pages 1877--1901.

\bibitem[{Cai et~al.(2024)Cai, Li, Geng, Peng, Lee, Chen, and Dao}]{medusa}
Tianle Cai, Yuhong Li, Zhengyang Geng, Hongwu Peng, Jason~D. Lee, Deming Chen,
  and Tri Dao. 2024.
\newblock Medusa: Simple {LLM} inference acceleration framework with multiple
  decoding heads.
\newblock In \emph{Proceedings of ICML}.

\bibitem[{Chen et~al.(2023)Chen, Borgeaud, Irving, Lespiau, Sifre, and
  Jumper}]{deepmindspec}
Charlie Chen, Sebastian Borgeaud, Geoffrey Irving, Jean-Baptiste Lespiau,
  Laurent Sifre, and John Jumper. 2023.
\newblock Accelerating large language model decoding with speculative sampling.
\newblock \emph{arXiv preprint arXiv:2302.01318}.

\bibitem[{Chen et~al.(2021)Chen, Tworek, Jun, Yuan, Pinto, Kaplan, Edwards,
  Burda, Joseph, Brockman et~al.}]{humaneval}
Mark Chen, Jerry Tworek, Heewoo Jun, Qiming Yuan, Henrique Ponde de~Oliveira
  Pinto, Jared Kaplan, Harri Edwards, Yuri Burda, Nicholas Joseph, Greg
  Brockman, et~al. 2021.
\newblock Evaluating large language models trained on code.
\newblock \emph{arXiv preprint arXiv:2107.03374}.

\bibitem[{Chen et~al.(2016)Chen, Xu, Zhang, and Guestrin}]{chen2016training}
Tianqi Chen, Bing Xu, Chiyuan Zhang, and Carlos Guestrin. 2016.
\newblock Training deep nets with sublinear memory cost.
\newblock \emph{arXiv preprint arXiv:1604.06174}.

\bibitem[{Cobbe et~al.(2021)Cobbe, Kosaraju, Bavarian, Chen, Jun, Kaiser,
  Plappert, Tworek, Hilton, Nakano, Hesse, and Schulman}]{gsm8k}
Karl Cobbe, Vineet Kosaraju, Mohammad Bavarian, Mark Chen, Heewoo Jun, Lukasz
  Kaiser, Matthias Plappert, Jerry Tworek, Jacob Hilton, Reiichiro Nakano,
  Christopher Hesse, and John Schulman. 2021.
\newblock Training verifiers to solve math word problems.
\newblock \emph{arXiv preprint arXiv:2110.14168}.

\bibitem[{Dao(2023)}]{flashattention-2}
Tri Dao. 2023.
\newblock Flashattention-2: Faster attention with better parallelism and work
  partitioning.
\newblock \emph{arXiv preprint arXiv:2307.08691}.

\bibitem[{Dong et~al.(2024)Dong, Ruan, Cai, Lai, Xu, Zhao, and
  Chen}]{dong2024xgrammar}
Yixin Dong, Charlie~F Ruan, Yaxing Cai, Ruihang Lai, Ziyi Xu, Yilong Zhao, and
  Tianqi Chen. 2024.
\newblock Xgrammar: Flexible and efficient structured generation engine for
  large language models.
\newblock \emph{arXiv preprint arXiv:2411.15100}.

\bibitem[{Dubey et~al.(2024)Dubey, Jauhri, Pandey, Kadian, Al-Dahle, Letman,
  Mathur, Schelten, Yang, Fan et~al.}]{llama3}
Abhimanyu Dubey, Abhinav Jauhri, Abhinav Pandey, Abhishek Kadian, Ahmad
  Al-Dahle, Aiesha Letman, Akhil Mathur, Alan Schelten, Amy Yang, Angela Fan,
  et~al. 2024.
\newblock The llama 3 herd of models.
\newblock \emph{arXiv preprint arXiv:2407.21783}.

\bibitem[{Fu et~al.(2024)Fu, Bailis, Stoica, and Zhang}]{fu2024break}
Yichao Fu, Peter Bailis, Ion Stoica, and Hao Zhang. 2024.
\newblock Break the sequential dependency of {LLM} inference using lookahead
  decoding.
\newblock In \emph{Proceedings of ICML}.

\bibitem[{Guo et~al.(2025)Guo, Yang, Zhang, Song, Zhang, Xu, Zhu, Ma, Wang, Bi
  et~al.}]{deepseekr1}
Daya Guo, Dejian Yang, Haowei Zhang, Junxiao Song, Ruoyu Zhang, Runxin Xu,
  Qihao Zhu, Shirong Ma, Peiyi Wang, Xiao Bi, et~al. 2025.
\newblock Deepseek-r1: Incentivizing reasoning capability in llms via
  reinforcement learning.
\newblock \emph{arXiv preprint arXiv:2501.12948}.

\bibitem[{Han et~al.(2021)Han, Zhang, Ding, Gu, Liu, Huo, Qiu, Yao, Zhang,
  Zhang et~al.}]{han2021pre}
Xu~Han, Zhengyan Zhang, Ning Ding, Yuxian Gu, Xiao Liu, Yuqi Huo, Jiezhong Qiu,
  Yuan Yao, Ao~Zhang, Liang Zhang, et~al. 2021.
\newblock Pre-trained models: Past, present and future.
\newblock \emph{AI Open}, 2:225--250.

\bibitem[{He et~al.(2024)He, Zhong, Cai, Lee, and He}]{rest}
Zhenyu He, Zexuan Zhong, Tianle Cai, Jason Lee, and Di~He. 2024.
\newblock {REST}: Retrieval-based speculative decoding.
\newblock In \emph{Proceedings of NAACL}, pages 1582--1595.

\bibitem[{Hokamp and Liu(2017)}]{hokamp2017lexically}
Chris Hokamp and Qun Liu. 2017.
\newblock Lexically constrained decoding for sequence generation using grid
  beam search.
\newblock In \emph{Proceedings of ACL}, pages 1535--1546.

\bibitem[{Joulin et~al.(2017)Joulin, Ciss{\'e}, Grangier, J{\'e}gou
  et~al.}]{joulin2017efficient}
Armand Joulin, Moustapha Ciss{\'e}, David Grangier, Herv{\'e} J{\'e}gou, et~al.
  2017.
\newblock Efficient softmax approximation for gpus.
\newblock In \emph{Proceedings of ICML}, pages 1302--1310. PMLR.

\bibitem[{Kwiatkowski et~al.(2019)Kwiatkowski, Palomaki, Redfield, Collins,
  Parikh, Alberti, Epstein, Polosukhin, Devlin, Lee, Toutanova, Jones, Kelcey,
  Chang, Dai, Uszkoreit, Le, and Petrov}]{natural_questions}
Tom Kwiatkowski, Jennimaria Palomaki, Olivia Redfield, Michael Collins, Ankur
  Parikh, Chris Alberti, Danielle Epstein, Illia Polosukhin, Jacob Devlin,
  Kenton Lee, Kristina Toutanova, Llion Jones, Matthew Kelcey, Ming-Wei Chang,
  Andrew~M. Dai, Jakob Uszkoreit, Quoc Le, and Slav Petrov. 2019.
\newblock Natural questions: A benchmark for question answering research.
\newblock \emph{TACL}, 7:453--466.

\bibitem[{Kwon et~al.(2023)Kwon, Li, Zhuang, Sheng, Zheng, Yu, Gonzalez, Zhang,
  and Stoica}]{kwon2023efficient}
Woosuk Kwon, Zhuohan Li, Siyuan Zhuang, Ying Sheng, Lianmin Zheng, Cody~Hao Yu,
  Joseph Gonzalez, Hao Zhang, and Ion Stoica. 2023.
\newblock Efficient memory management for large language model serving with
  pagedattention.
\newblock In \emph{Proceedings of SOSP}, pages 611--626.

\bibitem[{Leviathan et~al.(2023)Leviathan, Kalman, and Matias}]{googlespec}
Yaniv Leviathan, Matan Kalman, and Yossi Matias. 2023.
\newblock Fast inference from transformers via speculative decoding.
\newblock In \emph{Proceedings of ICML}, pages 19274--19286.

\bibitem[{Li et~al.(2024{\natexlab{a}})Li, Xu, Huang, Chen, Li, Liu, Lian, Pan,
  Ding, Zhou et~al.}]{li2024large}
Jinhao Li, Jiaming Xu, Shan Huang, Yonghua Chen, Wen Li, Jun Liu, Yaoxiu Lian,
  Jiayi Pan, Li~Ding, Hao Zhou, et~al. 2024{\natexlab{a}}.
\newblock Large language model inference acceleration: A comprehensive hardware
  perspective.
\newblock \emph{arXiv preprint arXiv:2410.04466}.

\bibitem[{Li et~al.(2024{\natexlab{b}})Li, Wei, Zhang, and Zhang}]{eagle2}
Yuhui Li, Fangyun Wei, Chao Zhang, and Hongyang Zhang. 2024{\natexlab{b}}.
\newblock Eagle-2: Faster inference of language models with dynamic draft
  trees.
\newblock In \emph{Proceedings of EMNLP}, pages 7421--7432.

\bibitem[{Liu et~al.(2024)Liu, Feng, Xue, Wang, Wu, Lu, Zhao, Deng, Zhang, Ruan
  et~al.}]{deepseekv3}
Aixin Liu, Bei Feng, Bing Xue, Bingxuan Wang, Bochao Wu, Chengda Lu, Chenggang
  Zhao, Chengqi Deng, Chenyu Zhang, Chong Ruan, et~al. 2024.
\newblock Deepseek-v3 technical report.
\newblock \emph{arXiv preprint arXiv:2412.19437}.

\bibitem[{Luo et~al.(2024)Luo, Zhao, Chen, Chen, and Anandkumar}]{luo2024mini}
Cheng Luo, Jiawei Zhao, Zhuoming Chen, Beidi Chen, and Anima Anandkumar. 2024.
\newblock Mini-sequence transformer: Optimizing intermediate memory for long
  sequences training.
\newblock \emph{arXiv preprint arXiv:2407.15892}.

\bibitem[{Miao et~al.(2024)Miao, Oliaro, Zhang, Cheng, Wang, Zhang, Wong, Zhu,
  Yang, Shi et~al.}]{specinfer}
Xupeng Miao, Gabriele Oliaro, Zhihao Zhang, Xinhao Cheng, Zeyu Wang, Zhengxin
  Zhang, Rae Ying~Yee Wong, Alan Zhu, Lijie Yang, Xiaoxiang Shi, et~al. 2024.
\newblock Specinfer: Accelerating large language model serving with tree-based
  speculative inference and verification.
\newblock In \emph{Proceedings of ASPLOS}, pages 932--949.

\bibitem[{Nallapati et~al.(2016)Nallapati, Zhou, dos Santos,
  Gu{\ensuremath{\dot{}}}l{\c{c}}ehre, and Xiang}]{cnn/daily_mail}
Ramesh Nallapati, Bowen Zhou, Cicero dos Santos, {\c{C}}a{\u{g}}lar
  Gu{\ensuremath{\dot{}}}l{\c{c}}ehre, and Bing Xiang. 2016.
\newblock Abstractive text summarization using sequence-to-sequence {RNN}s and
  beyond.
\newblock In \emph{Proceedings of CoNLL}, pages 280--290.

\bibitem[{OpenAI(2022)}]{chatgpt}
TB~OpenAI. 2022.
\newblock Chatgpt: Optimizing language models for dialogue.
\newblock \emph{OpenAI}.

\bibitem[{Paszke et~al.(2019)Paszke, Gross, Massa, Lerer, Bradbury, Chanan,
  Killeen, Lin, Gimelshein, Antiga et~al.}]{paszke2019pytorch}
Adam Paszke, Sam Gross, Francisco Massa, Adam Lerer, James Bradbury, Gregory
  Chanan, Trevor Killeen, Zeming Lin, Natalia Gimelshein, Luca Antiga, et~al.
  2019.
\newblock Pytorch: An imperative style, high-performance deep learning library.
\newblock \emph{Proceedings of NeurIPS}, 32.

\bibitem[{Qiu et~al.(2020)Qiu, Sun, Xu, Shao, Dai, and Huang}]{qiu2020pre}
Xipeng Qiu, Tianxiang Sun, Yige Xu, Yunfan Shao, Ning Dai, and Xuanjing Huang.
  2020.
\newblock Pre-trained models for natural language processing: A survey.
\newblock \emph{Science China Technological Sciences}, 63(10):1872--1897.

\bibitem[{Saxena(2023)}]{pld}
Apoorv Saxena. 2023.
\newblock \href {https://github.com/apoorvumang/prompt-lookup-decoding/}
  {Prompt lookup decoding}.

\bibitem[{ShareGPT(2023)}]{sharegpt2023}
ShareGPT. 2023.
\newblock \href
  {https://huggingface.co/datasets/anon8231489123/ShareGPT_Vicuna_unfiltered}
  {Sharegpt}.

\bibitem[{Soboleva et~al.(2023)Soboleva, Al-Khateeb, Myers, Steeves, Hestness,
  and Dey}]{cerebras2023slimpajama}
Daria Soboleva, Faisal Al-Khateeb, Robert Myers, Jacob~R Steeves, Joel
  Hestness, and Nolan Dey. 2023.
\newblock {SlimPajama: A 627B token cleaned and deduplicated version of
  RedPajama}.

\bibitem[{Sun et~al.(2024)Sun, Chen, Yang, Tian, and Chen}]{triforce}
Hanshi Sun, Zhuoming Chen, Xinyu Yang, Yuandong Tian, and Beidi Chen. 2024.
\newblock Triforce: Lossless acceleration of long sequence generation with
  hierarchical speculative decoding.
\newblock In \emph{Proceedings of COLM}.

\bibitem[{Takase et~al.(2024)Takase, Ri, Kiyono, and Kato}]{takase2024large}
Sho Takase, Ryokan Ri, Shun Kiyono, and Takuya Kato. 2024.
\newblock Large vocabulary size improves large language models.
\newblock \emph{arXiv preprint arXiv:2406.16508}.

\bibitem[{Tao et~al.(2024)Tao, Liu, Dou, Muennighoff, Wan, Luo, Lin, and
  Wong}]{tao2024scaling}
Chaofan Tao, Qian Liu, Longxu Dou, Niklas Muennighoff, Zhongwei Wan, Ping Luo,
  Min Lin, and Ngai Wong. 2024.
\newblock Scaling laws with vocabulary: Larger models deserve larger
  vocabularies.
\newblock In \emph{Proceedings of NeurIPS}, volume~37, pages 114147--114179.

\bibitem[{Touvron et~al.(2023)Touvron, Martin, Stone, Albert, Almahairi,
  Babaei, Bashlykov, Batra, Bhargava, Bhosale et~al.}]{llama2}
Hugo Touvron, Louis Martin, Kevin Stone, Peter Albert, Amjad Almahairi, Yasmine
  Babaei, Nikolay Bashlykov, Soumya Batra, Prajjwal Bhargava, Shruti Bhosale,
  et~al. 2023.
\newblock Llama 2: Open foundation and fine-tuned chat models.
\newblock \emph{arXiv preprint arXiv:2307.09288}.

\bibitem[{Wang et~al.(2024)Wang, Su, Li, Xia, Ye, Duan, Wang, and
  Zhang}]{wang2024opt}
Jikai Wang, Yi~Su, Juntao Li, Qingrong Xia, Zi~Ye, Xinyu Duan, Zhefeng Wang,
  and Min Zhang. 2024.
\newblock Opt-tree: Speculative decoding with adaptive draft tree structure.
\newblock \emph{arXiv preprint arXiv:2406.17276}.

\bibitem[{Wijmans et~al.(2024)Wijmans, Huval, Hertzberg, Koltun, and
  Kr{\"a}henb{\"u}hl}]{wijmans2024cut}
Erik Wijmans, Brody Huval, Alexander Hertzberg, Vladlen Koltun, and Philipp
  Kr{\"a}henb{\"u}hl. 2024.
\newblock Cut your losses in large-vocabulary language models.
\newblock \emph{arXiv preprint arXiv:2411.09009}.

\bibitem[{Xia et~al.(2023)Xia, Ge, Wang, Chen, Wei, and Sui}]{microsoftspec}
Heming Xia, Tao Ge, Peiyi Wang, Si-Qing Chen, Furu Wei, and Zhifang Sui. 2023.
\newblock Speculative decoding: Exploiting speculative execution for
  accelerating seq2seq generation.
\newblock In \emph{Proceedings of EMNLP}, pages 3909--3925.

\bibitem[{Xia et~al.(2024)Xia, Yang, Dong, Wang, Li, Ge, Liu, Li, and
  Sui}]{specbench}
Heming Xia, Zhe Yang, Qingxiu Dong, Peiyi Wang, Yongqi Li, Tao Ge, Tianyu Liu,
  Wenjie Li, and Zhifang Sui. 2024.
\newblock Unlocking efficiency in large language model inference: A
  comprehensive survey of speculative decoding.
\newblock In \emph{Findings of the ACL}, pages 7655--7671.

\bibitem[{Xu and McAuley(2023)}]{xu2023survey}
Canwen Xu and Julian McAuley. 2023.
\newblock A survey on model compression and acceleration for pretrained
  language models.
\newblock In \emph{Proceedings of AAAI}, volume~37, pages 10566--10575.

\bibitem[{Yang et~al.(2024{\natexlab{a}})Yang, Yang, Hui, Zheng, Yu, Zhou, Li,
  Li, Liu, Huang, Dong et~al.}]{qwen2}
An~Yang, Baosong Yang, Binyuan Hui, Bo~Zheng, Bowen Yu, Chang Zhou, Chengpeng
  Li, Chengyuan Li, Dayiheng Liu, Fei Huang, Guanting Dong, et~al.
  2024{\natexlab{a}}.
\newblock Qwen2 technical report.

\bibitem[{Yang et~al.(2024{\natexlab{b}})Yang, Yang, Zhang, Hui, Zheng, Yu, Li,
  Liu, Huang, Wei et~al.}]{qwen25}
An~Yang, Baosong Yang, Beichen Zhang, Binyuan Hui, Bo~Zheng, Bowen Yu,
  Chengyuan Li, Dayiheng Liu, Fei Huang, Haoran Wei, et~al. 2024{\natexlab{b}}.
\newblock Qwen2.5 technical report.
\newblock \emph{arXiv preprint arXiv:2412.15115}.

\bibitem[{Yang et~al.(2023)Yang, Ge, Wang, Jiao, Jiang, Yang, Majumder, and
  Wei}]{llma}
Nan Yang, Tao Ge, Liang Wang, Binxing Jiao, Daxin Jiang, Linjun Yang, Rangan
  Majumder, and Furu Wei. 2023.
\newblock Inference with reference: Lossless acceleration of large language
  models.
\newblock \emph{arXiv preprint arXiv:2304.04487}.

\bibitem[{Yao et~al.(2025)Yao, Chen, Zhang, You, Yuan, Wang, and
  Lin}]{yao2025deft}
Jinwei Yao, Kaiqi Chen, Kexun Zhang, Jiaxuan You, Binhang Yuan, Zeke Wang, and
  Tao Lin. 2025.
\newblock De{FT}: Decoding with flash tree-attention for efficient
  tree-structured {LLM} inference.
\newblock In \emph{Proceedings of ICLR}.

\bibitem[{Zhang et~al.(2025)Zhang, Wang, Huang, and Xu}]{zhang2025learning}
Lefan Zhang, Xiaodan Wang, Yanhua Huang, and Ruiwen Xu. 2025.
\newblock Learning harmonized representations for speculative sampling.
\newblock In \emph{Proceedings of ICLR}.

\bibitem[{Zhang et~al.(2024)Zhang, Wang, Ma, Zhu, Chen, Lan, and
  Yu}]{zhang2024adaeagle}
Situo Zhang, Hankun Wang, Da~Ma, Zichen Zhu, Lu~Chen, Kunyao Lan, and Kai Yu.
  2024.
\newblock Adaeagle: Optimizing speculative decoding via explicit modeling of
  adaptive draft structures.
\newblock \emph{arXiv preprint arXiv:2412.18910}.

\bibitem[{Zhao et~al.(2023)Zhao, Zhou, Li, Tang, Wang, Hou, Min, Zhang, Zhang,
  Dong et~al.}]{zhao2023survey}
Wayne~Xin Zhao, Kun Zhou, Junyi Li, Tianyi Tang, Xiaolei Wang, Yupeng Hou,
  Yingqian Min, Beichen Zhang, Junjie Zhang, Zican Dong, et~al. 2023.
\newblock A survey of large language models.
\newblock \emph{arXiv preprint arXiv:2303.18223}.

\bibitem[{Zhao et~al.(2024)Zhao, Huang, Han, Xu, Xiao, Zhang, Fang, Zhang, Liu,
  and Sun}]{zhao-etal-2024-ouroboros}
Weilin Zhao, Yuxiang Huang, Xu~Han, Wang Xu, Chaojun Xiao, Xinrong Zhang, Yewei
  Fang, Kaihuo Zhang, Zhiyuan Liu, and Maosong Sun. 2024.
\newblock Ouroboros: Generating longer drafts phrase by phrase for faster
  speculative decoding.
\newblock In \emph{Proceedings of EMNLP}, pages 13378--13393.

\bibitem[{Zheng et~al.(2023)Zheng, Chiang, Sheng, Zhuang, Wu, Zhuang, Lin, Li,
  Li, Xing, Zhang, Gonzalez, and Stoica}]{mtbench}
Lianmin Zheng, Wei-Lin Chiang, Ying Sheng, Siyuan Zhuang, Zhanghao Wu, Yonghao
  Zhuang, Zi~Lin, Zhuohan Li, Dacheng Li, Eric Xing, Hao Zhang, Joseph~E
  Gonzalez, and Ion Stoica. 2023.
\newblock Judging llm-as-a-judge with mt-bench and chatbot arena.
\newblock In \emph{Proceedings of CoNLL}, volume~36, pages 46595--46623.

\bibitem[{Zheng et~al.(2024)Zheng, Yin, Xie, Sun, Huang, Yu, Cao, Kozyrakis,
  Stoica, Gonzalez et~al.}]{zheng2024sglang}
Lianmin Zheng, Liangsheng Yin, Zhiqiang Xie, Chuyue Sun, Jeff Huang, Cody~Hao
  Yu, Shiyi Cao, Christos Kozyrakis, Ion Stoica, Joseph~E Gonzalez, et~al.
  2024.
\newblock Sglang: Efficient execution of structured language model programs.
\newblock \emph{arXiv preprint arXiv:2312.07104}.

\bibitem[{Zhou et~al.()Zhou, Lyu, Rawat, Menon, Rostamizadeh, Kumar, Kagy, and
  Agarwal}]{distillspec}
Yongchao Zhou, Kaifeng Lyu, Ankit~Singh Rawat, Aditya~Krishna Menon, Afshin
  Rostamizadeh, Sanjiv Kumar, Jean-Fran{\c{c}}ois Kagy, and Rishabh Agarwal.
\newblock Distillspec: Improving speculative decoding via knowledge
  distillation.
\newblock In \emph{Proceedings of ICLR}.

\bibitem[{Zipf(1950)}]{zipf-law}
George~Kingsley Zipf. 1950.
\newblock Human behavior and the principle of least effort: An introduction to
  human ecology.
\newblock \emph{Language}, 26:394.

\end{thebibliography}

\newpage
\appendix
\section{Additional Results}

\subsection{Qwen-2-7B Performance}
\label{appendix:qwen2-accept-length}

\begin{table*}[t]
\centering
\scalebox{0.85}{
\begin{tabular}{l|ccccccc|c}
\toprule
Configuration & \textbf{MT.} & \textbf{Conv.} & \textbf{RAG} & \textbf{Math} & \textbf{QA} & \textbf{Summ.} & \textbf{Code} & \textbf{Average} \\
\midrule
Full Vocab (152k) & 2.90 & 4.06 & 3.65 & 4.31 & 3.27 & 3.74 & 4.22 & 3.74 (100\%) \\
\midrule
+FR 64k (ShareGPT) & 2.86 & 3.98 & 3.65 & 4.22 & 3.23 & 3.67 & 4.17 & 3.68 (98.6\%)  \\
+FR 32k (ShareGPT) & 2.76 & 3.90 & 3.42 & 4.10 & 3.24 & 3.39 & 3.98 & 3.54 (94.8\%) \\
+FR 16k (ShareGPT) & 2.62 & 3.64 & 3.20 & 3.85 & 2.99 & 3.08 & 3.71 & 3.30 (88.3\%) \\
+FR 8k (ShareGPT) & 2.45 & 3.39 & 3.01 & 3.60 & 2.48 & 2.81 & 3.41 & 3.02 (80.9\%) \\
\midrule
+FR 64k (SlimPajama) & 2.90 & 3.97 & 3.64 & 4.29 & 3.28 & 3.73 & 3.98 & 3.69 (98.6\%) \\
+FR 32k (SlimPajama) & 2.83 & 3.73 & 3.53 & 4.20 & 3.39 & 3.58 & 3.71 & 3.57 (95.4\%) \\
+FR 16k (SlimPajama) & 2.67 & 3.50 & 3.33 & 3.95 & 3.25 & 3.35 & 3.40 & 3.35 (89.7\%) \\
+FR 8k (SlimPajama) & 2.60 & 3.28 & 3.12 & 3.65 & 2.91 & 3.04 & 3.10 & 3.10 (83.0\%) \\
\bottomrule
\end{tabular}
}
\caption{Average accepted length for Qwen-2-7B on under different \name~configurations.}
\label{tab:acceptance_qwen}
\end{table*}

\begin{table*}[t]
\centering
\scalebox{0.85}{
\begin{tabular}{l|ccccccc|c}
\toprule
Configuration & \textbf{MT.} & \textbf{Conv.} & \textbf{RAG} & \textbf{Math} & \textbf{QA} & \textbf{Summ.} & \textbf{Code} & \textbf{Average} \\
\midrule
Full Vocab (128k)  & 2.49 & 2.96 & 2.80 & 3.08 & 2.69 & 2.62 & 3.04 & 2.81 (100\%) \\
\midrule
+FR 64k (ShareGPT) & 2.43 & 2.93 & 2.75 & 3.05 & 2.67 & 2.58 & 2.98 & 2.77 (98.6\%)  \\ 
+FR 32k (ShareGPT) & 2.39 & 2.90 & 2.65 & 2.98 & 2.54 & 2.51 & 2.85 & 2.69 (95.7\%)  \\ 
+FR 16k (ShareGPT) & 2.34 & 2.78 & 2.56 & 2.88 & 2.42 & 2.42 & 2.75 & 2.59 (92.3\%)  \\ 
+FR 8k (ShareGPT) & 2.25 & 2.66 & 2.44 & 2.76 & 2.35 & 2.31 & 2.65 & 2.49 (88.6\%) \\ 
\midrule
+FR 64k (SlimPajama) & 2.47 & 2.92 & 2.78 & 3.07 & 2.68 & 2.61 & 2.88 & 2.77 (98.7\%)  \\ 
+FR 32k (SlimPajama) & 2.43 & 2.82 & 2.69 & 3.04 & 2.58 & 2.57 & 2.70 & 2.69 (95.8\%)  \\ 
+FR 16k (SlimPajama) & 2.38 & 2.72 & 2.62 & 2.91 & 2.51 & 2.50 & 2.58 & 2.60 (92.6\%)  \\ 
+FR 8k (SlimPajama) & 2.30 & 2.58 & 2.50 & 2.80 & 2.40 & 2.39 & 2.43 & 2.49 (88.5\%) \\ 
\bottomrule
\end{tabular}
}
\caption{Average accepted length for Llama-3.2-1B on under different \name~configurations.}
\label{tab:acceptance_llama3_1b}
\end{table*}

\begin{table*}[t]
\centering
\scalebox{0.85}{
\begin{tabular}{l|ccccccc|c}
\toprule
Method & \textbf{MT.} & \textbf{Conv.} & \textbf{RAG} & \textbf{Math} & \textbf{QA} & \textbf{Summ.} & \textbf{Code} & \textbf{Average} \\
\midrule
Vanilla & 259.83  & 255.89  & 220.25  & 263.34  & 260.13  & 248.15  & 256.64  & 252.03 (1.00$\times$)\\ 
\midrule
EAGLE-2 & 306.04  & 358.37  & 266.84  & 372.37  & 305.52  & 294.82  & 360.60  & 323.51 (1.28$\times$)\\ 
\midrule
+FR 64k & 349.12  & 406.14  & 297.62  & 427.14  & 350.08  & 338.81  & 390.78  & 365.67 (1.45$\times$) \\ 
+FR 32k & 378.90  & 428.75  & 317.68  & 467.53  & 378.39  & 363.70  & 395.95  & 390.13 (1.55$\times$)\\ 
\textbf{+FR 16k} & \textbf{394.81}  & \textbf{443.00}  & \textbf{326.75}  & \textbf{476.47}  & \textbf{394.47}  & \textbf{375.70}  & \textbf{402.07}  & \textbf{401.90 (1.59$\times$)} \\ 
+FR 8k & 386.97  & 428.94  & 319.83  & 462.98  & 382.75  & 363.50  & 392.13  & 391.01 (1.55$\times$) \\ 
\bottomrule
\end{tabular}
}
\caption{Decoding speed (token/s) of \name~and other baselines on Llama-3.2-1B under our implementation using temperature=0 and SlimPajama token-frequency statistics. The numbers in parentheses (1.59$\times$) indicate the speedup compared to the baseline (Vanilla).}
\label{tab:speed_llama3_1b_temp0}
\end{table*}

Following the settings in Section~\ref{sec:accept_length}, we investigated the impact of \name~on draft model's accepted length in the Qwen-2-7B model, which has a different vocabulary. The results in Table~\ref{tab:acceptance_qwen} show that the decrease ratio in acceptance length across various configuration settings in Qwen-2-7B is similar to or even less than that observed in Llama-3-8B, indicating the effectiveness of our method on various LLMs.

\subsection{Llama-3.2-1B Performance}
\label{appendix:llama3-1b}

Following the settings in Section~\ref{sec:accept_length} and Section~\ref{sec:exp_speed},
we conducted accept length and speed experiments on the Llama-3.2-1B model using a single 3090 GPU. Given the smaller size of the model, we adjusted the drafting depth of Eagle-2 to 3 and set the total number of draft tokens to 30.

The average acceptance length obtained from the experiments is presented in Table~\ref{tab:acceptance_llama3_1b}, while the speedup ratio in our implemented framework is shown in Table~\ref{tab:speed_llama3_1b_temp0}.
Results show that \name~achieves an extra 1.24$\times$ speedup over the state-of-the-art EAGLE-2. The speedup is even higher than the experimental results for Llama-3-8B. This demonstrates the effectiveness of \name, particularly on less powerful hardware and smaller LLMs.

\begin{figure}[h]
    \centering
    \includegraphics[width=0.44\textwidth]{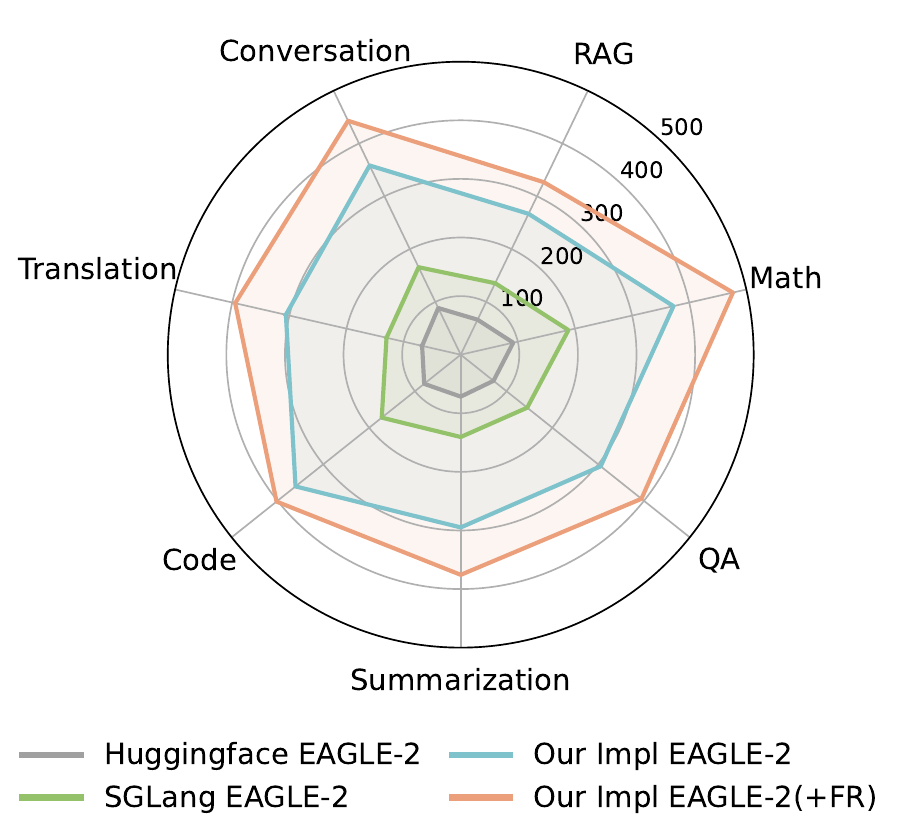}
    \caption{Decoding speed (token/s) of \name~and EAGLE-2 for Llama-3.2-1B under different implementation framework.}
    \label{fig:speed_compare_llama3_1b}
\end{figure}

Speed comparison with other frameworks is illustrated in Figure~\ref{fig:speed_compare_llama3_1b}. The overall speedup ratio of \name~was 5.24$\times$ and 2.61$\times$ compared with Huggingface and SGLang, respectively.

\end{document}